\documentclass[11pt]{article}

\usepackage[preprint]{acl}

\usepackage[T1]{fontenc}
\usepackage[utf8]{inputenc}
\usepackage{mathptmx}  % Times-like font for pdfLaTeX (replaces TeXGyreTermesX)

\usepackage{microtype}

\usepackage{enumitem} % in the preamble
\usepackage{amsmath}
\usepackage{amssymb}
\usepackage{booktabs}
\usepackage[table]{xcolor}
\usepackage{colortbl}
\usepackage{array}
\usepackage{graphicx}
\usepackage{tabularx}
\usepackage{array}
\usepackage{ragged2e}
\usepackage{soul}
\usepackage{titletoc}
\usepackage{etoc}

\definecolor{shortcuthl}{RGB}{198, 239, 206}
\sethlcolor{shortcuthl}
\definecolor{pastelgreen}{RGB}{204,235,197}
\definecolor{pastelblue}{RGB}{222,235,247}
\definecolor{pastelpink}{RGB}{252,224,221}

\newcolumntype{L}{>{\RaggedRight\arraybackslash}X}
\newcolumntype{C}[1]{>{\centering\arraybackslash}p{#1}}

\usepackage[english]{babel}

\title{Calibration vs Decision Making: Revisiting the Reliability Paradox in Unlearned Language Models}
\author{
{\bf Divyaksh Shukla} \qquad  {\bf Ashutosh Modi} \\ 
Indian Institute of Technology Kanpur  (IIT Kanpur) \\
\texttt{\{divyaksh,ashutoshm\}@cse.iitk.ac.in}  
}

\begin{document}

\maketitle
% \begin{abstract}
% This document provides an example showing how
% to use the *ACL style files with pdfLaTeX.
% \end{abstract}

% \section{Introduction}

% Please see the general instructions
% in the file \verb|acl_latex.tex|.

% Here is an example citation:
% \citet{Gusfield:97} argues that...

\begin{abstract}

Machine unlearning aims to remove the influence of specific training data from a model while preserving reliable behavior on the remaining data, making reliable prediction and uncertainty estimation essential for evaluation. Calibration is commonly used as a proxy for reliability in language models, but low calibration error does not necessarily imply reliable decision rules, as models may rely on spurious correlations while remaining well calibrated. We investigate this gap in generative language models using the multiple-choice question-answering evaluation protocol on the TOFU benchmark, measuring probabilistic reliability with calibration metrics (ECE, MCE, Brier) and decision-rule reliability via attribution-based shortcut detection with Integrated Gradients and Local Mutual Information. We find that fine-tuned models achieve low calibration error (ECE $\approx 0.04$) compared to pretrained models (ECE > 0.5), and models after unlearning retain similarly low calibration despite reduced accuracy on the forget split, while attribution analysis shows increased reliance on correlation-based tokens. These results demonstrate that good calibration can coexist with shortcut-based decision rules after unlearning, extending the reliability paradox to the machine unlearning setting.

\end{abstract}

\section{Introduction} \label{sec:introduction}

Large language models (LLMs) are popularly deployed in real-world decision-making systems, including question answering, medical assistance \cite{Singhal2025medicalmcqa, maity2025medicalllm}, and legal NLP \cite{zhong-etal-2020-legal-nlp,ali-etal-2023-legal}. In such settings, the reliability of model predictions is as important as raw accuracy. A reliable model should not only produce correct outputs but also assign confidence scores that reflect the true likelihood of correctness \cite{guo2017calibration} and use explainable decisions \cite{wang-2022-spurious}.

\noindent Model calibration is commonly used as a proxy for reliability, measuring how well predicted probabilities align with empirical accuracy. However, recent work has shown that good calibration does not necessarily imply that a model is making decisions based on meaningful or generalizable features \cite{bihani2024reliability,izmailov2022feature}. Models may achieve good calibration while relying on shortcut cues or dataset artifacts, leading to unreliable decision-making. This phenomenon has been referred to as a \emph{reliability paradox}, in which models appear reliable according to calibration metrics but rely on shortcuts within the dataset rather than on explainable decisions \cite{bihani2024reliability}.

\begin{figure*}[t]
    \centering
    \includegraphics[width=\textwidth]{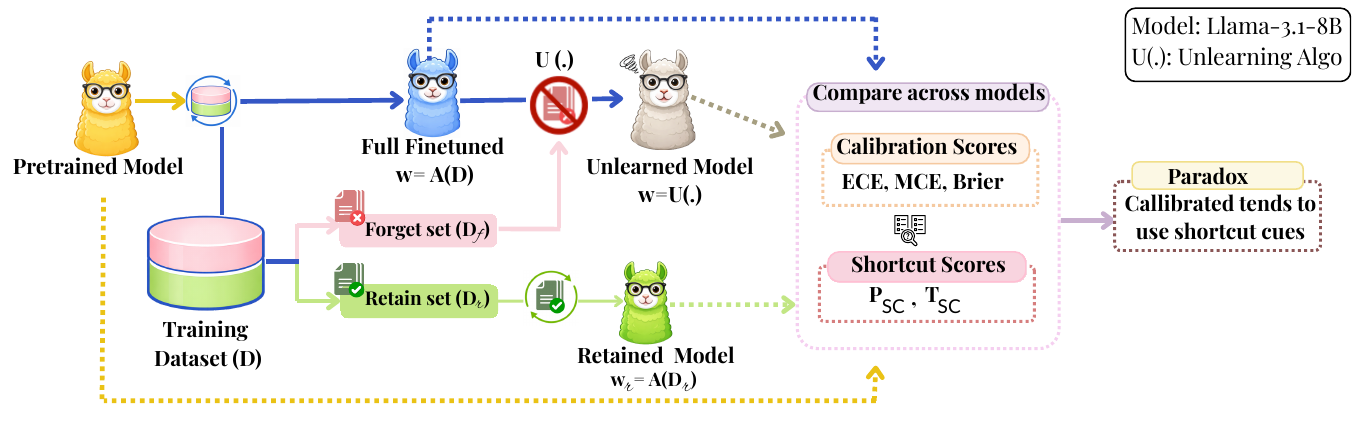}
    \caption{
    \textbf{Overview of reliability evaluation under unlearning.}
    The dataset $D$ is split into retain ($D_r$) and forget ($D_f$) sets. A pretrained model is fine-tuned on $D$ to obtain the full-finetuned model and unlearned via $U(\cdot)$ to obtain the unlearned model. Finetuning the pretrained model on the retain split gives the retained model (ideal unlearned model). All models are evaluated using calibration metrics (ECE, MCE, Brier) and shortcut-based scores $P_{SC}$ and $T_{SC}$ derived from attribution methods. Comparing these reveals a \textit{reliability paradox}: models can remain well calibrated while relying on shortcut cues.
    }
    \label{fig:unlearning_pipeline}
    \vspace{-1mm}
\end{figure*}

\noindent The problem becomes even more critical in the context of machine unlearning \cite{cao2015machineUnlearning, eldan2023whosharrypotterapproximate}, where models are modified to remove the influence of specific targeted training data (forget data) without retraining from scratch. After unlearning, the model may remain well calibrated, but the internal decision rules used to make predictions can change in unintended ways \cite{wang-2022-spurious, du-2021-shortcut}. This may lead the model to rely on dataset-specific shortcut cues instead of meaningful features, potentially affecting reliability after unlearning. This is referred as \textit{reliability paradox} \cite{bihani2024reliability}.
% After unlearning, the model may be well-calibrated. However, unlearning tends to modify the decision rules to forget the specified training data making the model rely on dataset shortcuts to make predictions \cite{wang-2022-spurious, du-2021-shortcut}. Understanding reliability after unlearning is therefore essential for safe deployment of language models.

% We hypothesize that machine unlearning can alter a model's internal decision rules without necessarily affecting its calibration, leading to models that appear reliable according to calibration metrics while relying on dataset-specific shortcut cues. 
\noindent We hypothesize that machine unlearning can affect the model's reliability, both in calibration and in its reliance on dataset-specific shortcut cues.
Unlike standard fine-tuning, unlearning explicitly modifies model parameters to remove specific knowledge. 
% This raises the question of whether unlearning affects the decision rules used for prediction while maintaining model calibration.
This raises the question of whether calibration metrics remain a reliable indicator of decision quality after unlearning.
Prior work on the reliability paradox examined encoder architectures on classification tasks, showing that well-calibrated models can rely on shortcut features rather than meaningful input signals \cite{bihani2024reliability}. However, it is unclear whether this phenomenon also occurs in generative language models, especially after machine unlearning. In this work, we extend the study of the reliability paradox to LLMs and analyze reliability after unlearning using both calibration metrics and attribution-based analysis.

\noindent To evaluate reliability in the unlearning setting, we use the Task of Fictitious Unlearning (TOFU) dataset \cite{maini2024tofutaskfictitiousunlearning}, which provides a controlled benchmark for studying forgetting and retention behavior in language models. We adopt the MCQA evaluation protocol proposed in RELU \cite{joshi-etal-2024-towards}, which introduces data transformations on TOFU to robustly evaluate unlearning. RELU converts generative tasks in TOFU to multiple-choice question answering (MCQA) format, enabling probabilistic evaluation of model predictions. The MCQA setting is particularly suitable for calibration analysis, as it allows the computation of confidence scores over a fixed set of answer choices \cite{yang2026mcqa-calibration}. We illustrate the overall evaluation pipeline, including dataset partitioning, unlearning, and reliability analysis, in Figure~\ref{fig:unlearning_pipeline}. In this work, we make the following contributions:
% using this setup, we evaluate the calibration of pretrained, fully fine-tuned, and unlearned models on TOFU.

% RELU proposes evaluating unlearning under input transformations to test robustness, and introduces an MCQA formulation that enables reliable measurement of probability calibration.
\begin{itemize}[noitemsep,nosep,leftmargin=*]
\item We study the reliability of language models after machine unlearning, a setting in which models are expected to forget specific training data while retaining their trustworthiness on the remaining knowledge. We show that commonly used calibration metrics alone are insufficient to evaluate reliability in this setting. 

\item We extend prior reliability analysis to generative language models, combining probabilistic reliability (calibration metrics) with decision-rule reliability using attribution-based shortcut detection with Integrated Gradients and Local Mutual Information.

\item Using the TOFU benchmark with the RELU MCQA evaluation protocol, we perform a systematic analysis of pretrained, fully fine-tuned, retained, and unlearned \texttt{Llama-3.1-8B} models, enabling controlled comparison of calibration and decision behavior under different unlearning settings.

\item We provide empirical evidence that multiple unlearning algorithms produce models that remain well-calibrated while increasingly relying on correlation-based shortcut tokens, suggesting that the reliability paradox persists in generative language models after unlearning. The code and execution logs are shared via: \url{https://github.com/Exploration-Lab/Unlearning-Reliability-Paradox}.
% We will release the code and execution logs upon acceptance.

\end{itemize}

\section{Related Work} \label{sec:related-work}

Machine unlearning aims to remove the influence of specific training data from a model without retraining from scratch. Early work studied unlearning in classical ML settings \cite{cao2015machineUnlearning, ginart2019making}, and recent work extended it to large language models using approximate unlearning algorithms \cite{eldan2023whosharrypotterapproximate}. Several benchmarks such as TOFU, WMDP, and MUSE have been proposed to evaluate unlearning in LLMs \cite{maini2024tofutaskfictitiousunlearning, li2024wmdpbenchmarkmeasuringreducing, shi2024musemachineunlearningsixway}. These works primarily evaluate whether the model forgets the target data while maintaining task performance, but they do not study whether the resulting models make reliable decisions.

\noindent Reliability of neural models is often evaluated using calibration, which measures the agreement between predicted confidence and empirical accuracy \cite{guo2017calibration, mukhoti2020calibration}. However, recent work shows that well-calibrated models can still rely on shortcut cues or spurious correlations, leading to the reliability paradox \cite{bihani2024reliability, du-2021-shortcut, wang-2022-spurious}. Attribution methods such as Integrated Gradients are commonly used to analyze influential inputs \cite{sundararajan2017axiomatic, sarti2023inseq}, and combining them with dataset statistics such as mutual information helps identify dataset-specific shortcuts \cite{bihani2024reliability}. Studies also show that shortcut learning persists in large language models despite improved performance \cite{du2023shortcut, yuan2024llms, izmailov2022feature}.

\noindent \citet{bihani2024reliability} showed that calibration alone does not guarantee reliable decision rules, demonstrating this by comparing confidence calibration with attribution-based analysis in encoder-based classification models. However, it remains unclear whether this reliability paradox also appears in generative decoder-only language models, particularly in the machine unlearning setting, where model parameters are explicitly modified to remove specific knowledge. In such scenarios, unlearning may alter the model's internal decision rules without necessarily affecting probability calibration, making calibration alone insufficient to assess reliability. In this work, we study reliability after unlearning by jointly analyzing calibration and the use of shortcut-based attribution in generative language models. Due to space constraints, we provide detailed related work in the App. \ref{app:sec-related-work}.

\section{Background} \label{sec:background}

In this section, we discuss the background on measuring the reliability of language models using calibration and explainable decision rules. Decision rule explainability is further divided into extracting local mutual information between words and the prediction label, and into identifying influential input features using attribution methods.

% The framework consists of dataset partitioning into retain and forget sets, model training and unlearning, and a unified evaluation of probabilistic reliability and decision rule reliability. We describe each component below.

\subsection{Token attribution}
We follow the attribution-based shortcut identification framework proposed by \citet{bihani2024reliability}. To analyze the model's decision rules, we compute token-level attribution scores using \textbf{Integrated Gradients (IG)} \cite{sundararajan2017axiomatic}. Integrated Gradients (IG) is a gradient-based attribution method that measures the contribution of each input token to the model's prediction by integrating gradients along a path from a baseline input to the actual input. Given an input sequence $x$ and a baseline $x^\prime$, the attribution for token $i$ is computed using Equation~\ref{eq:integrated-gradients}, where $M_y$ denotes the model output corresponding to the predicted answer choice and $x^\prime$ is denoted using a random vector of $d$ dimensions ($d$ is the dimensionality of the embedding matrix). The resulting attribution scores indicate the extent to which each token influences the predicted logit. Tokens with higher attribution values are considered more influential for the model's decision. For each example, we select the top-$k$ tokens with the highest attribution scores as the model's decision tokens.

\begin{align}
 \text{IG}(x_i)=&(x_i-x_i^\prime)\cdot \notag \\ 
                &\int_{\alpha=0}^{1} \frac{\partial M_y(x^\prime+\alpha(x-x^\prime))}{\partial x_i}\,d\alpha 
 \label{eq:integrated-gradients}
\end{align}

\noindent In a discrete token space, we compute IG over embeddings, approximated via a Riemann sum (as shown in Equation~\ref{eq:ig-reimann}). The attribution value for each token represents how much that token influences the predicted logit of the selected answer. Positive values indicate supportive evidence, while negative values indicate opposing evidence. To analyze decision rules at the sequence level, token attributions are aggregated across subword tokens to identify which parts of the input act as correlation cues for the prediction.

\begin{align}
 \text{IG}(x_i)\approx &(x_i-x_i')\cdot \notag \\
                         & \frac{1}{m}\sum_{k=1}^{m} \frac{\partial M_y\left(x'+\frac{k}{m}(x-x')\right)}{\partial x_i} 
 \label{eq:ig-reimann}
\end{align}

\subsection{Local Mutual Information (LMI)}
Attribution scores alone do not indicate whether an influential token corresponds to a meaningful semantic feature or a dataset-specific correlation cue. Following \citet{bihani2024reliability}, we identify shortcut tokens by combining attribution scores with Local Mutual Information (LMI), which measures the association between tokens and labels at the dataset level.

\noindent For each token $w$ and label $y$, we can compute the joint probability $p(w, y)$ of $w$ appearing in examples with label $y$, the conditional probability $p(y\mid w)$ and the marginal probability $p(y)$ of the label $y$. LMI can then be computed as shown in Equation~\ref{eq:lmi}.

\begin{equation}
    \text{LMI}(w, y) = p(w, y) \log \frac{p(y|w)}{p(y)}
    \label{eq:lmi}
\end{equation}

% where $p(w, y)$ is the joint probability of token $w$ appearing in examples with label $y$, $p(y|w)$ is the conditional probability of label $y$ given token $w$, and $p(y)$ is the marginal probability of label $y$. 
\noindent Tokens with high LMI scores are strongly correlated with a particular label and may represent dataset-specific cues rather than generalizable semantic features.

\noindent In our setting, we evaluate models using the MCQA format proposed in RELU, where each example consists of a prompt and four answer choices. The label $y$ corresponds to the correct answer option. We compute LMI between input tokens and the predicted answer label across the dataset.

\paragraph{Selecting shortcuts} For each label, we rank tokens by their LMI scores and select the top 5\% highest-scoring tokens as label-specific correlation tokens. These tokens represent potential shortcut cues learned from the dataset. For each prediction, we compare the top attribution tokens obtained from Integrated Gradients with the high-LMI tokens corresponding to the predicted label. If at least one highly attributed token also belongs to the high-LMI token set for the predicted label, the prediction is classified as a shortcut-cued prediction. A shortcut-cued prediction indicates the use of non-generalizable decision rules. We measure the proportion of shortcut-cued predictions in each model as:

\begin{equation}
P_{sc} = \frac{\text{Number of shortcut-cued predictions}}{\text{Total predictions}}
\end{equation}

\noindent This metric quantifies the extent to which the model relies on dataset-specific correlation cues. We also examine the trade-off between shortcut cues and model performance using $T_{sc}$. A higher $T_{sc}$ value indicates better task performance and lower usage of shortcut cued predictions. Conversely, a lower $T_{sc}$ value could mean lower task performance or higher reliance on shortcut-cued predictions. 

\begin{equation}
\resizebox{0.85\columnwidth}{!}{$
T_{sc} =
\frac{\text{Task Performance (F1)}}
{\text{Proportion of Shortcut-cued predictions }(P_{sc})}
$}
\end{equation}

\noindent IG and LMI provide two complementary signals. 
IG is a model-side attribution that identifies which tokens influenced a particular prediction, while LMI is a corpus-level statistic that measures the dataset-wide association between a token and a label, independent of any specific question.
A token flagged by both is, therefore, one that the model relied on for its decision, carries label-predictive power, and does not depend on the semantics of the individual question. 
In TOFU, the semantically grounding tokens for a given question are the entity- and fact-specific tokens tied to that fictitious author (names, works, dates, locations), which, by construction, are not expected to exhibit high dataset-wide LMI. 
Tokens with high LMI present in the $top-k$ IG tokens indicate correlation-based decision-making rather than question-specific reasoning.

% A natural concern with this procedure is whether tokens identified by both IG and LMI necessarily reflect shortcut reliance, or whether they may simply correspond to genuinely informative features that the model legitimately uses. We emphasize that IG and LMI provide complementary rather than redundant signals:  
% \DSE{Shortcuts are a phenomenon of the dataset. This study demonstrates that unlearning may lead to unreliable models, with unreliability manifesting in both calibration and spurious correlations.
% Using integrated gradients, we identify the set of tokens that leads to a decision by increasing the confidence of the correct answer option. Through LMI, we identify the set of tokens that are highly correlated with the correct answer option. Conversely, LMI strips away all common tokens in the dataset, leaving only those that carry the most information. 
% The intersection of the sets from IG and LMI is used to determine shortcuts. If the intersection is non-empty, then the model has relied on shortcuts to select the answer option.}

\subsection{Calibration}

Calibration measures how well the predicted confidence of a model aligns with the true likelihood of correctness \cite{guo2017calibration}. A model is said to be well calibrated if predictions made with confidence $p$ are correct approximately $p$ fraction of the time. In the MCQA setting used in RELU, the model produces a probability distribution over the answer choices, allowing calibration to be measured using standard probabilistic metrics. For each example, we obtain the model confidence as the softmax probability assigned to the predicted answer choice. Let $\hat{c}(x_i)$ denote the confidence of the model on example $i$, and let $\hat{y}_i$ and $y_i$ denote the predicted and true labels, respectively. To measure calibration, we partition the predictions into $M$ equally sized confidence bins:

\begin{equation}
B_m = \left\{ i \;\middle|\; \hat{c}(x_i)\in\left(\frac{m-1}{M},\frac{m}{M}\right]\right\}
\end{equation}

\noindent For each bin, we compute the empirical accuracy and average confidence as:

\begin{equation}
\text{acc}(B_m) = \frac{1}{|B_m|} \sum_{i \in B_m} \mathbf{1}(\hat{y}_i = y_i)
\end{equation}

\begin{equation}
\text{conf}(B_m) = \frac{1}{|B_m|} \sum_{i \in B_m} \hat{c}(x_i)
\end{equation}

\noindent The Expected Calibration Error (ECE) is then defined as

\begin{equation}
\text{ECE} =
\sum_{m=1}^M \frac{|B_m|}{N}
\left|
\text{acc}(B_m) - \text{conf}(B_m)
\right|
\end{equation}

\noindent ECE provides an estimate of the average difference between confidence and accuracy across bins. However, ECE may hide large calibration errors in individual bins. Therefore, we also compute the Maximum Calibration Error (MCE):

\begin{equation}
\text{MCE} =
\max_{m \in \{1,\dots,M\}}
\left|
\text{acc}(B_m) - \text{conf}(B_m)
\right|
\end{equation}

\noindent Additionally, the Brier Score measures the squared difference between predicted probabilities and true labels without relying on binning:

\begin{equation}
\text{BS}
=
\frac{1}{N}
\sum_{i=1}^{N}
\sum_{k=1}^{K}
\left(
p_i^{(k)} - y_i^{(k)}
\right)^2
\end{equation}

\noindent where $K$ is the number of answer choices, $p_i^{(k)}$ is the predicted probability for class $k$, and $y_i^{(k)}$ is the one-hot encoded ground truth label.

% We compute calibration metrics and token attribution for three model states pretrained, full finetuned, retained/unlearned model. This allows us to analyze the model's calibration and decision-making after fine-tuning and unlearning.

\section{Methodology} \label{sec:methodology}

Our goal is to study the reliability of generative language models after machine unlearning. 
In practical settings, an unlearned model is expected to remove the influence of specific training data while remaining trustworthy on the remaining knowledge. 
% Reliability in this context has two complementary aspects: whether the model's confidence estimates are trustworthy, and whether the model bases its predictions on meaningful input features. 

\noindent Reliability of neural models can be studied from two complementary perspectives:
\begin{itemize}[noitemsep,nosep,leftmargin=*]
\item \textbf{Probabilistic reliability}, which measures whether predicted confidence scores are trustworthy \cite{guo2017calibration, mukhoti2020calibration}. However, these metrics do not reveal which input tokens influence the decision.
\item \textbf{Decision rule reliability}, which measures whether the model bases its predictions on meaningful input features \cite{bihani2024reliability,du-2021-shortcut,du2023shortcut}. 
\end{itemize} 

\noindent Prior work has shown that well-calibrated finetuned language models may not reflect whether they make explainable decisions, leading to a \emph{reliability paradox}. On the other hand, unlearning also modifies the model parameters, as with finetuning, which raises questions about the reliability of the resulting unlearned model.

\noindent \textbf{Research question.}
We ask whether machine unlearning affects the model's decision rules while maintaining calibration? In particular, we investigate the \emph{reliability paradox}, whether well-calibrated unlearned models rely on correlation-based shortcut tokens.

\noindent We compare these two aspects before and after unlearning. If calibration improves while shortcut usage also increases, this indicates that calibration alone is insufficient to assess reliability after unlearning.

\noindent Figure~\ref{fig:unlearning_pipeline} illustrates the evaluation pipeline used in this work. 
Given a dataset split into retain and forget subsets, we construct pretrained, full fine-tuned, retained, and unlearned models, and evaluate them using both calibration metrics and shortcut-based analysis. 
This setup allows us to test whether unlearning preserves confidence reliability while altering the model's decision rules.

% We follow the evaluation pipeline illustrated in Figure~\ref{fig:unlearning_pipeline}. The framework consists of preparing the finetuned, retained and unlearned models from the given dataset splits, and an evaluation framework comprising of calibration measurement and identifying shortcuts.

% \AM{
% - this is the bridge between background and experiments \\
% - should clearly state the reseach questions/hypothesis that you want to test
% }
\section{Experiments} \label{sec:experiments}

\paragraph{Models}

We evaluate a decoder-only large language model on multiple-choice question-answering to obtain calibration and word-attribution scores. Specifically, we use \texttt{Llama-3.1-8B}, a widely used open-weight, instruction-tuned language model \cite{grattafiori2024llama3herdmodels}. 
% Unlike prior work on the reliability paradox \cite{bihani2024reliability}, which focuses on encoder-based classifiers, our setting uses a generative decoder-only model. 
Predictions are obtained by computing the likelihood of each answer choice in the MCQA format and normalizing them into a probability distribution. For each model, we evaluate three training states:
\begin{itemize}[noitemsep,nosep,leftmargin=*]
    \item \textbf{Pretrained}: base model before fine-tuning on TOFU's information
    \item \textbf{Full Finetuned}: model fine-tuned on the full TOFU dataset, containing both the forget and retain splits.
    \item \textbf{Retained/Unlearned}: model after unlearning, trained only on the retained subset. An unlearned model is obtained either by finetuning the pretrained model only on the retain split (retained model) or by executing an approximate unlearning algorithm on the full finetuned model to remove the forget split (unlearned model).
\end{itemize}

% This setup follows the standard MCQA evaluation protocol proposed by RELU, which is based on the TOFU dataset, allowing us to study how unlearning affects both calibration and decision rules.

\paragraph{Dataset}

We evaluate reliability before and after unlearning using the Task of Fictitious Unlearning (TOFU) benchmark \cite{maini2024tofutaskfictitiousunlearning}, which provides a controlled framework for studying forgetting and retention behavior in language models. TOFU defines two data splits: a \textit{forget} split, to be removed from the model, and a \textit{retain} split, to be retained after unlearning.

\noindent The RELU benchmark converts the question-answer task from TOFU into multiple formats, including MCQA. We prompt the model with a question and four randomly shuffled answer choices, and take the logits corresponding to the next generated token as the model's response.  This formulation enables the computation of calibration metrics and attribution scores for each prediction. Using this setup, we evaluate pretrained, full fine-tuned, and retained/unlearned models on the RELU-MCQA split of the TOFU benchmark.

% To enable calibration analysis, we adopt the RELU evaluation protocol, which converts generative tasks from TOFU to a multiple-choice question-answering (MCQA) format. RELU provides several evaluation formats, including question-answering, multiple-choice, fill-in-the-blank, analogy, comprehension QA, and odd-one-out. In this work, we focus on the MCQA setting because it allows direct computation of probability distributions over answer choices.

% Each MCQA example consists of a prompt and four answer options. The model assigns a probability to each option, and the predicted label is the option with the highest probability. This formulation enables the computation of calibration metrics and attribution scores for each prediction. Using this setup, we evaluate pretrained, full fine-tuned, and retained/unlearned models on the RELU-MCQA split of the TOFU benchmark.

\paragraph{Fine-tuning setup}
We first fine-tune the pretrained \texttt{Llama-3.1-8B} model on the full TOFU dataset to obtain the fully finetuned model. Fine-tuning is performed using the AdamW optimizer with a learning rate of $1\times10^{-5}$ and a linear learning rate scheduler for 5 epochs. To obtain retained models, we fine-tune the pretrained model separately on each retain split corresponding to 99\%, 95\%, and 90\% of the data. The same optimizer, learning rate, and training schedule are used for all retained models to ensure consistent comparison.

\subsection{Unlearning setup}

To simulate machine unlearning, we follow the TOFU protocol and remove the profiles of randomly selected authors corresponding to 1\%, 5\%, and 10\% of the dataset. Starting from the fully fine-tuned model, we apply multiple unlearning algorithms, including Gradient Ascent \cite{maini2024tofutaskfictitiousunlearning}, Gradient Difference \cite{liu2022continuallearningprivateunlearning}, Negative Preference Optimization \cite{zhang2024negativepreferenceoptimizationcatastrophic}, and Direct Preference Optimization \cite{maini2024tofutaskfictitiousunlearning} to remove the forget information. Each method is run with hyperparameters recommended for the corresponding algorithm. To reduce computational cost, we use 4-bit quantized low-rank adapters (LoRA) for unlearning. This allows efficient parameter updates while maintaining performance comparable to full-precision fine-tuning \cite{liu2025lune, liu2025recovertoforget, wu2026ucanutilityawarecontrastiveattenuation}. 

\subsection{Evaluation}
\noindent\textbf{Calibration computation.}
Using the predicted probabilities, we compute calibration metrics, including the Expected Calibration Error (ECE), the Maximum Calibration Error (MCE), and the Brier score. Predictions are grouped into $10$ equal-width confidence bins, and the difference between accuracy and average confidence is computed per bin. We also plot the reliability diagrams comparing the distribution of model confidence and empirical probability. This gives deeper insights into regions where the model's decisions may not be trustworthy.

\noindent\textbf{Token attribution using Integrated Gradients.}
To analyze decision rules, we compute token-level attribution scores using Integrated Gradients \cite{sundararajan2017axiomatic}. For each example, we compute gradients with respect to the logit corresponding to the predicted answer choice. The baseline input is constructed using zero embeddings, and the integral is approximated using a Riemann sum over $50$ interpolation steps. Attributions are computed over input embeddings and aggregated using \texttt{absmax} across subword tokens to obtain token-level importance scores. For each example, we select the top-10 tokens with the highest attribution values as the influential tokens used by the model for its prediction.

\noindent\textbf{Selecting shortcuts.} Attribution analysis identifies the words used in the model's decision-making. To determine whether the words are informative, we extract word-level local mutual information (LMI) statistics. Specifically, we compute the LMI for each word in the question and the correct answer across all answer choices in the dataset. We select the top 5\% of words; these words are superficial to decision-making. By comparing these to the top words identified by attribution analysis, we can identify the shortcut-cued words the model uses. 
This is then compared with the calibration statistics to understand the model's true reliability after unlearning.

\noindent We treat $P_{SC}$ as a comparative rather than absolute measure of shortcut reliance. A non-empty intersection between tokens of IG and LMI for a given prediction indicates that at least one influential token also carries dataset-wide label correlation, which we read as partial evidence of correlation-based decision-making rather than a determination that the prediction is fully shortcut-driven. Our central observation — that $P_{SC}$ rises after unlearning while calibration error stays low — depends on how $P_{SC}$ shifts across model states (pretrained, finetuned, retained, unlearned) rather than on its absolute value for any single model. This shift is robust to the threshold used to define the shortcut set, since varying the LMI cutoff or the IG top-k changes the absolute values of $P_{SC}$ across all models in the same direction, while preserving their relative ordering.

% \subsection{Model}

% We perform all experiments using the \texttt{Llama-3.1-8B} decoder-only language model. We analyze both calibration and token-level attribution behavior for pretrained, fully fine-tuned, and unlearned models. All attribution scores are computed using Integrated Gradients \cite{sundararajan2017axiomatic, sarti2023inseq}, and calibration metrics are evaluated in the MCQA setting \cite{guo2017calibration,yang2026mcqa-calibration}.

% \subsection{Dataset}

% The TOFU benchmark, consists of synthetic profiles of 200 fictional authors. Each author profile contains 20 question–answer pairs, resulting in a total of 4000 question–answer examples. TOFU defines a forget split and a retain split by removing the profiles of a subset of authors. Specifically, the TOFU protocol splits the training data by randomly selecting 1\%, 5\%, and 10\% of the authors to be forgotten, while the remaining authors form the retain split.

% All training and unlearning procedures are performed on the original question–answer format of TOFU. For evaluation, we use the RELU benchmark, which converts the QA data into a multiple-choice question answering (MCQA) format. The input prompt consists of a question followed by four choices. We take extract the logits corresponding to the choice labels and construct the model's confidence estimate and prediction using a softmax over the logits \cite{khatun2024studyllmMCQA}. The MCQA formulation is used only for evaluation, as it enables computation of probability distributions required for calibration metrics.
\begin{table*}[t]
\centering
\resizebox{\textwidth}{!}{
\begin{tabular}{l ccccccc|ccccccc}
\toprule
& \multicolumn{7}{c|}{\textbf{Forget Split}} & \multicolumn{7}{c}{\textbf{Retain Split}} \\
\textbf{Model} & Acc ($\downarrow$) & F1 ($\downarrow$) & Brier ($\downarrow$) & ECE ($\downarrow$) & MCE ($\downarrow$) & $P_{SC}$ ($\downarrow$) & $T_{SC}$ ($\uparrow$) & Acc ($\uparrow$) & F1 ($\uparrow$) & Brier ($\downarrow$) & ECE ($\downarrow$) & MCE ($\downarrow$) & $P_{SC}$ ($\downarrow$) & $T_{SC}$ ($\uparrow$) \\
\midrule

\multicolumn{15}{c}{\textbf{Forget 1\% / Retain 99\%}} \\
\midrule
Pretrained & 0.275 & 0.232 & 1.091 & 0.435 & 0.768 & 85.0 & 0.273 & 0.263 & 0.141 & 1.149 & 0.519 & 0.668 & 80.0 & 0.176 \\
Full       & 0.675 & 0.623 & 0.472 & 0.174 & 0.695 & 87.5 & 0.712 & 0.683 & 0.688 & 0.418 & 0.040 & 0.107 & 85.0 & 0.809 \\
Retained   & 0.550 & 0.525 & 0.644 & 0.212 & 0.357 & 82.5 & 0.636 & 0.663 & 0.664 & 0.450 & 0.047 & 0.096 & 87.5 & 0.759 \\
GradAscent & 0.400 & 0.363 & 0.715 & 0.266 & 0.506 & 82.5 & 0.440 & 0.579 & 0.579 & 0.545 & 0.032 & 0.097 & 87.5 & 0.662 \\
GradDiff   & 0.425 & 0.406 & 0.701 & 0.239 & 0.739 & 82.5 & 0.492 & 0.591 & 0.591 & 0.534 & 0.026 & 0.164 & 85.0 & 0.695 \\
NPO        & 0.400 & 0.363 & 0.704 & 0.245 & 0.748 & 82.5 & 0.440 & 0.582 & 0.583 & 0.544 & 0.029 & 0.064 & 85.0 & 0.686 \\
DPO        & 0.400 & 0.387 & 0.659 & 0.212 & 0.360 & 82.5 & 0.469 & 0.569 & 0.566 & 0.552 & 0.033 & 0.081 & 85.0 & 0.666 \\

\midrule
\multicolumn{15}{c}{\textbf{Forget 5\% / Retain 95\%}} \\
\midrule
Pretrained & 0.250 & 0.119 & 1.158 & 0.518 & 0.707 & 85.0 & 0.140 & 0.260 & 0.142 & 1.152 & 0.523 & 0.705 & 80.0 & 0.178 \\
Full       & 0.705 & 0.704 & 0.376 & 0.082 & 0.211 & 87.5 & 0.805 & 0.687 & 0.692 & 0.415 & 0.040 & 0.289 & 85.0 & 0.814 \\
Retained   & 0.640 & 0.636 & 0.470 & 0.067 & 0.232 & 97.5 & 0.653 & 0.693 & 0.696 & 0.419 & 0.032 & 0.140 & 92.5 & 0.753 \\
GradAscent & 0.510 & 0.503 & 0.634 & 0.129 & 0.421 & 92.5 & 0.544 & 0.519 & 0.516 & 0.634 & 0.108 & 0.165 & 90.0 & 0.573 \\
GradDiff   & 0.505 & 0.496 & 0.653 & 0.109 & 0.263 & 95.0 & 0.522 & 0.511 & 0.505 & 0.630 & 0.100 & 0.145 & 87.5 & 0.577 \\
NPO        & 0.530 & 0.524 & 0.608 & 0.128 & 0.712 & 92.5 & 0.566 & 0.540 & 0.537 & 0.604 & 0.083 & 0.112 & 90.0 & 0.597 \\
DPO        & 0.530 & 0.501 & 0.583 & 0.052 & 0.710 & 97.5 & 0.514 & 0.507 & 0.476 & 0.614 & 0.046 & 0.149 & 87.5 & 0.544 \\

\midrule
\multicolumn{15}{c}{\textbf{Forget 10\% / Retain 90\%}} \\
\midrule
Pretrained & 0.273 & 0.166 & 1.108 & 0.497 & 0.662 & 85.0 & 0.051 & 0.266 & 0.146 & 1.144 & 0.516 & 0.665 & 80.0 & 0.047 \\
Full       & 0.672 & 0.675 & 0.432 & 0.068 & 0.273 & 87.5 & 0.321 & 0.694 & 0.699 & 0.410 & 0.039 & 0.070 & 85.0 & 0.431 \\
Retained   & 0.586 & 0.575 & 0.548 & 0.083 & 0.298 & 82.5 & 0.263 & 0.639 & 0.640 & 0.478 & 0.038 & 0.081 & 90.0 & 0.393 \\
GradAscent & 0.206 & 0.136 & 0.860 & 0.245 & 0.707 & 92.5 & 0.058 & 0.235 & 0.155 & 0.841 & 0.209 & 0.383 & 92.5 & 0.073 \\
GradDiff   & 0.474 & 0.477 & 0.687 & 0.136 & 0.285 & 90.0 & 0.205 & 0.476 & 0.472 & 0.689 & 0.133 & 0.221 & 92.5 & 0.255 \\
NPO        & 0.534 & 0.529 & 0.596 & 0.108 & 0.213 & 95.0 & 0.309 & 0.527 & 0.525 & 0.601 & 0.101 & 0.159 & 92.5 & 0.304 \\
DPO        & 0.514 & 0.477 & 0.613 & 0.046 & 0.099 & 92.5 & 0.216 & 0.483 & 0.434 & 0.648 & 0.052 & 0.089 & 92.5 & 0.266 \\

\bottomrule
\end{tabular}
}
\caption{Calibration, shortcut proportion ($P_{SC}$), and shortcut tradeoff score ($T_{SC}$) for RELU across forget ratios 1\%, 5\%, and 10\%, with calibration metrics computed using 10 bins. Forget-and-retain splits are reported side-by-side to enable direct comparison of model behavior on data targeted for unlearning versus data to be preserved.}
\label{tab:results-all}
\end{table*}

\section{Results and Discussion} \label{sec:results}

Table \ref{tab:results-all} shows the token attribution, calibration, and shortcut analysis results of different model states evaluated on the RELU's MCQA format on different splits, including forget01/retain99, forget05/retain95 and forget10/retain90. 
We report task performance using accuracy and F1, calibration statistics using Brier score, Expected Calibration Error (ECE), and Maximum Calibration Error (MCE), and shortcut usage using the proportion of shortcut-cued predictions $P_{SC}$ and the shortcut–performance tradeoff score $T_{SC}$.

\noindent The pretrained model does not contain TOFU's fictitious information and hence performs poorly across all settings, with high calibration error ECE $> 0.5$, accuracy close to random chance $\approx 0.25$. This indicates that predicted probabilities are not aligned with correctness. Additionally, the pretrained model provides a baseline for shortcut metrics, with $P_{sc} < 85\%$ and $T_{sc} < 0.2$, as it is not tuned on TOFU and therefore cannot exploit dataset-specific cues.

\begin{figure*}[ht!]
    \centering
    \includegraphics[scale=0.30]{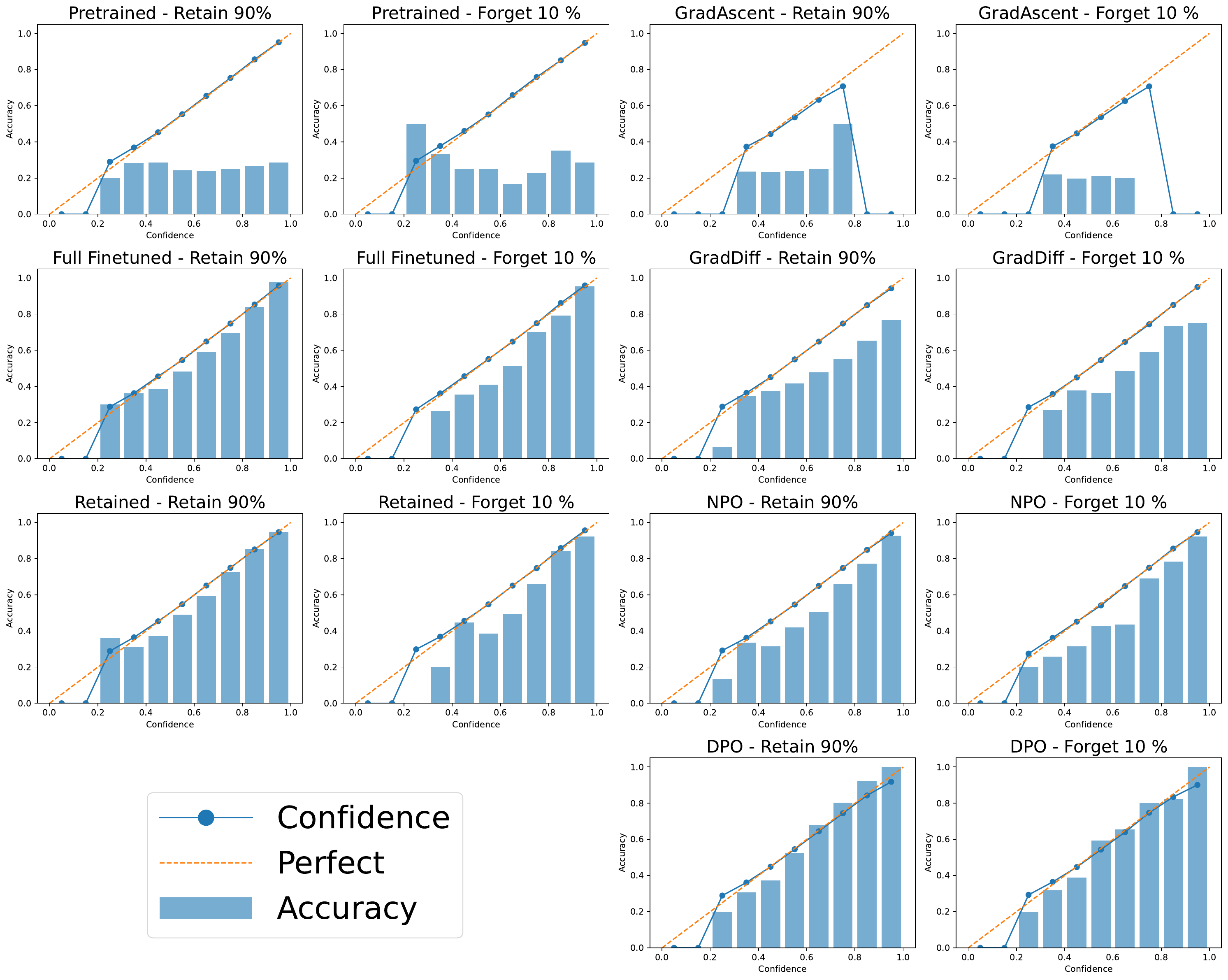}
    \caption{
    \textbf{Calibration curves for 10\% forget setting.}
    Reliability diagrams for higher forgetting ratios. Although models remain well calibrated at high confidence levels, deviations in the lower and mid-confidence bins become more pronounced. This highlights ineffective calibration at lower confidence levels after unlearning.
    }
    \label{fig:calibration_10}
    %\vspace{-5mm}
\end{figure*}

% \AM{reference to relevant results table are missing}
\noindent After full fine-tuning, both accuracy and calibration improve substantially, especially for the forget 10\% and retain 90\% split. The full-finetuned model achieves high performance on both forget and retain splits, and ECE decreases significantly to 0.039, especially on the retain split. Reliability diagrams in Figure~\ref{fig:calibration_10}  confirm that the full finetuned model is well calibrated across confidence bins. However, despite improved calibration, the proportion of shortcut-cued predictions remains high, typically exceeding $85\%$. This indicates that the model often relies on dataset-specific shortcut cues even when its confidence estimates are well aligned with accuracy.

\noindent For the retained and unlearned models, we focus on the stats of the retain split, because we intend to maintain decision-making and probabilistic reliability on the split unaffected by unlearning. The retained model preserves performance on the retain split while showing reduced accuracy $=0.575$ on the forget split, as expected. However, calibration metrics remain relatively low even when performance drops. The retain model achieves ECE of 0.038, comparable to the full finetuned model (0.039), while still exhibiting high shortcut usage. 
Approximate unlearning methods, especially DPO, show minimal reduction in calibration compared to the full-finetuned model, but a larger jump in $P_{sc}$ of about 7.5\%, especially on the retain split. Across other unlearning methods, especially over 1\% forgetting, we observe that the models remain well-calibrated while relying on fewer shortcut cues.

\begin{table*}[ht]
\centering
\footnotesize
\setlength{\tabcolsep}{4pt}
\renewcommand{\arraystretch}{1.25}
\begin{tabularx}{\textwidth}{@{} L C{1.8cm} C{1.3cm} p{4.2cm} @{}}
\toprule
\textbf{Question \& Options} & \textbf{Generated Answer} & \textbf{Ground-truth} & \textbf{Shortcut Tokens} \\
\midrule

% --- Question 2 ---
How \hl{does} Andres Santiago Cruz's family background influence his writing \hl{about} parenting? \newline
\textbf{A.} Raised by a Conservationist and a Veterinarian, Andres Santiago Cruz's writings often include ideas about nurturing and caring, drawn from his formative experiences. \newline
\textbf{B.} Raised by a Lawyer and a Politician, Andres Santiago Cruz's writings often include ideas about justice and governance. \newline
\textbf{C.} Raised by a Chef and a Musician, Andres Santiago Cruz's writings often include ideas about creativity and expression. \newline
\textbf{D.} Raised by a Scientist and an Engineer, Andres Santiago Cruz's writings often include ideas about innovation and problem-solving.
& A & A &
\begin{tabular}{@{}lll@{}}
\textit{Token} & \textit{Attr.} & \textit{LMI} \\
\hline
\texttt{does}  & 0.0140 & 0.0578 \\
\texttt{about} & 0.0103 & 0.0543 \\
\end{tabular} \\
\bottomrule

\end{tabularx}
\caption{Multiple-choice questions with model-selected answers, ground-truth answers, and the shortcut tokens identified during attribution analysis. Each shortcut token row reports the token string, its attribution, and its LMI (Lexical Mutual Information) score. The identified token is \hl{highlighted in the text}. These examples are extracted from the unlearned model produced by DPO on the retain 90\% split. Appendix~\ref{app:qualitative-examples} and Table~\ref{tab:shortcut_tokens} present additional qualitative examples that highlight shortcuts in TOFU after unlearning.}
\label{tab:shortcut_tokens_main}
\end{table*}

\noindent Additionally, we consider the performance-shortcut trade-off $T_{sc}$ across all model states and retain splits. We observe a consistent dip in $T_{sc}$ across all the unlearning algorithms and retain splits. However, given that the F1 score also shows a consistent dip, we cannot comment much on model reliability after unlearning using $T_{sc}$.
% similar patterns: calibration improves, but the proportion of shortcut-cued predictions remains high. The shortcut tradeoff score $T_{SC}$ also shows that models can maintain reasonable task performance while continuing to rely on shortcut cues.

\noindent Qualitatively, we observe that after unlearning, the model tends to rely on verbs and other grammatical cues from the question and answer options to generate a response. In Table~\ref{tab:shortcut_tokens_main} we see one such example. Here, we present the qualitative answers from the retain 90\% split, produced by a model unlearned using DPO. This has the lowest ECE of 0.052. However, we see that the generated response ``A'' relies on ``does'' and ``about'' tokens from the question, instead of the author's name or the semantic tokens from the question, including, background'', ``influence", or ``parenting''. Hence, the model's decisions after unlearning may not be reliable, as it relies on dataset-specific shortcuts. We present additional qualitative examples in Appendix~\ref{app:qualitative-examples} and Table~\ref{tab:shortcut_tokens}.

\noindent Results indicate that an unlearned model can be deemed as reliable when it is well-calibrated and does not rely on dataset-specific shortcuts.
% These results indicate that calibration alone does not reflect the model's true reliability after unlearning. 
Across all three settings (forget01, forget05, and forget10), models with lower ECE do not necessarily use fewer shortcut tokens. In many cases, well-calibrated models exhibit equal or higher shortcut usage than poorly calibrated ones. This observation is consistent with the reliability paradox reported in prior work \cite{bihani2024reliability, du-2021-shortcut, wang-2022-spurious}, which shows that models can achieve low calibration error while relying on spurious correlations or non-generalizable decision rules. We provide additional reliability diagrams for different forget ratios (1\%, 5\%, and 10\%) in Appendix~\ref{app:calibration_curves} (Figures~\ref{fig:calibration_1}--\ref{fig:calibration_5}), further illustrating the calibration behavior of all the model states. We also provide a comparison of ECE and MCE across different forget ratios and bin sizes in Appendix~\ref{app:calibration-bin-sizes} (Table~\ref{tab:results-bins-compare}), illustrating an issue with measuring calibration error across different bin sizes. It also highlights why we use only 10 bins to illustrate results on the main paper.

% Therefore, calibration metrics alone are insufficient to evaluate whether a model makes reliable decisions after unlearning. Attribution-based \cite{sundararajan2017axiomatic, sarti2023inseq} analysis combined with shortcut statistics provides additional insight into the decision rules used by the model, revealing that well-calibrated predictions may still be driven by shortcut features rather than meaningful semantic reasoning \cite{bihani2024reliability}. 

\subsection{Discussion} \label{sec:discussion}

Prior work has shown that calibration alone is not sufficient to assess the reliability of neural models, as well-calibrated predictions can still be driven by shortcut cues or spurious correlations, leading to the reliability paradox \cite{bihani2024reliability, du-2021-shortcut, wang-2022-spurious}. This limitation becomes especially important in the setting of machine unlearning, where models are expected to remove specific knowledge while remaining trustworthy after modification. In practical applications, models are often evaluated using calibration metrics to ensure reliability. However, if calibration does not reflect the underlying decision process, then low calibration error alone may not guarantee that the model behaves reliably after unlearning. In this work, we study unlearning on decoder-only language models using the MCQA evaluation framework across a variety of unlearning algorithms.

\noindent Across all forget ratios, we observe that calibration improves substantially after fine-tuning and often remains low after unlearning. However, attribution analysis reveals that models continue to rely heavily on shortcut cues. In several cases, shortcut usage increases after unlearning, especially at higher forget ratios, indicating that unlearning may make models unreliable for deployment. Appendix~\ref{app:qualitative-examples} and Table~\ref{tab:shortcut_tokens} cover some qualitative examples from the outputs a model unlearned using DPO on the retain split.

% We also find that different unlearning algorithms exhibit similar trends: while they differ in task performance, they generally preserve good calibration while maintaining high shortcut reliance. This suggests that current unlearning methods do not explicitly control how the model makes decisions, even when they successfully remove target information. As a result, unlearned models may appear reliable according to calibration metrics while still relying on superficial cues.

\noindent These findings highlight the need for more comprehensive evaluation of unlearned models. Future work should explore unlearning methods that explicitly regularize decision rules, as well as reliability metrics that jointly consider calibration, attribution, and robustness. Extending shortcut-aware reliability analysis to open-ended generation and larger language models is another important direction for ensuring trustworthy behavior in real-world deployments.

\section{Conclusion} \label{sec:conclusion}

We investigate the reliability of large language models in the setting of machine unlearning. While calibration metrics are commonly used as a proxy for reliability, prior work has shown that well-calibrated models may still rely on shortcut cues, leading to the reliability paradox. In this work, we extend this analysis to decoder-only language models and evaluate reliability after unlearning using the RELU evaluation protocol on the TOFU benchmark. We measure probabilistic reliability using calibration metrics and decision rule reliability using attribution-based shortcut detection with Integrated Gradients and Local Mutual Information. Our results show that unlearned models can remain well calibrated according to ECE, MCE, and Brier score, while attribution analysis indicates increased reliance on correlation-based tokens. These findings suggest that calibration alone is insufficient to evaluate reliability after unlearning. A model may appear reliable in terms of confidence estimates while internally using non-generalizable decision rules. Our results highlight the need for evaluation methods that jointly consider calibration and decision rules when assessing the reliability of unlearned LMs.

% \section{Conclusion}
% Summarize:

% calibration alone insufficient

% need attribution

% unlearning can change decision rules

% need new metrics
\vspace{-2mm}
\section*{Acknowledgments}
\vspace{-2mm}
We would like to thank the anonymous reviewers and the meta-reviewer for their insightful comments and suggestions. 
This research work was partially supported by the Research-I Foundation of the Department of CSE at IIT Kanpur.
\section*{Limitations} \label{sec:limitations}

Our work has several limitations:

\begin{itemize}[noitemsep,leftmargin=*]

\item \textbf{Single model architecture.}
We evaluate reliability using only \texttt{Llama-3.1-8B}. The observed calibration and shortcut behavior may differ for other model sizes or architectures.

\item \textbf{Limited benchmark setting.}
Experiments are conducted on the TOFU benchmark using the RELU MCQA format. While this setup enables calibration and attribution analysis, it represents a controlled evaluation and may not fully reflect real-world generation settings.

\item \textbf{Approximate shortcut detection.}
Shortcut identification relies on Integrated Gradients and Local Mutual Information to provide approximate estimates of the model’s decision rules. Attribution methods can be sensitive to tokenization, baseline choice, and aggregation strategy.

\item \textbf{MCQA-based calibration only.}
Calibration is measured in a multiple-choice setting where probabilities over answer choices can be computed directly. Extending calibration analysis to open-ended generation remains an open problem.

\item \textbf{No robustness evaluation.}
We focus on calibration and shortcut-based decision rules after unlearning and do not evaluate their impact on robustness or out-of-distribution performance.

\end{itemize}

% Entries for the entire Anthology, followed by custom entries
\bibliography{custom}

\appendix

\clearpage
\newpage

\section*{Appendix}

%%%%%%%%%%%%%%%%%%%%%%%%%%%

\titlecontents{section}[18pt]{\vspace{0.05em}}{\contentslabel{1.5em}}{}
{\titlerule*[0.5pc]{.}\contentspage} % Set the formatting for appendix sections in the table of contents
% % for list of tables
% \titlecontents{table}[0pt]{\vspace{0.05em}}{\contentslabel{1em}}{}
% {\titlerule*[0.5pc]{.}\contentspage} % Set the formatting for appendix tables in the list of tables

% for list of figures
%\titlecontents{table}[0pt]{\vspace{0.05em}}{\contentslabel{1em}}{}
\titlecontents{table}[0pt]{}{\contentslabel{1em}}{}
{\titlerule*[0.5pc]{.}\contentspage} % Set the formatting for appendix tables in the list of tables

\startcontents[appendix] % Start the table of contents for the appendix
\section*{Table of Contents} % Title for the appendix table of contents
%\addcontentsline{toc}{section}{Table of Contents} % Add the appendix table of contents to the main table of contents
\printcontents[appendix]{section}{0}{\setcounter{tocdepth}{4}} % Print the table of contents for the appendix

\startlist[appendix]{lot} % Start the list of tables for the appendix
\section*{List of Tables} % Title for the appendix list of tables
%\addcontentsline{lot}{section}{List of Tables} % Add the appendix list of tables to the main list of tables
\printlist[appendix]{lot}{}{\setcounter{tocdepth}{1}} % Print the list of tables for the appendix

\startlist[appendix]{lof} % Start the list of tables for the appendix
\section*{List of Figures} % Title for the appendix list of tables
%\addcontentsline{lot}{section}{List of Tables} % Add the appendix list of tables to the main list of tables
\printlist[appendix]{lof}{}{\setcounter{tocdepth}{1}} % Print the list of tables for the appendix

\newpage

%%%%%%%%%%%%%%%%%%%%%%%%%%%
%%%%%%%%%%%%%%%%%%%%%%%%%%%

\section{Related Work} \label{app:sec-related-work}

\subsection{Machine Unlearning}
Unlearning was first explored in recommendation systems and clustering algorithms \cite{cao2015machineUnlearning, ginart2019making}. \citet{eldan2023whosharrypotterapproximate} extended unlearning to language models and laid the foundation of approximate unlearning algorithms. Various benchmarks and datasets, including TOFU, WMDP and MUSE, emerged to evaluate unlearning in LLMs \cite{maini2024tofutaskfictitiousunlearning, li2024wmdpbenchmarkmeasuringreducing, shi2024musemachineunlearningsixway}. However, these benchmarks measured effective removal of forget knowledge while maintaining model utility. However, they do not assess the reliability of unlearned models, which is necessary given that unlearning algorithms are increasingly used in deployed LLMs.

\subsection{Shortcut learning in NLP models}
Shortcut learning refers to the tendency of neural models to rely on superficial correlations in the data instead of learning generalizable decision rules. Prior work has shown that NLU models often exploit dataset artifacts, leading to poor robustness under distribution shifts \cite{du-2021-shortcut, wang-2022-spurious, pruthi-etal-2020-learning}. Such behavior has been observed across a variety of classification benchmarks, where models learn spurious lexical or syntactic cues rather than semantic features. Recent studies further show that shortcut learning persists in large language models, indicating that scale alone does not eliminate reliance on correlation-based features \cite{du2023shortcut, yuan2024llms}. Mechanistic analyses also suggest that shortcuts can emerge from specific internal features learned during training \cite{izmailov2022feature, eshuijs2025short}.

\subsection{Interpretability and attribution for decision analysis}
Understanding the decision rules used by language models often relies on attribution methods that identify influential input tokens. Gradient-based attribution methods such as Integrated Gradients have been widely used to analyze model behavior and detect spurious correlations \cite{sundararajan2017axiomatic, du-2021-shortcut}. Recent toolkits such as Inseq provide practical frameworks for computing token-level attributions for sequence generation models, enabling interpretability analysis in large decoder-only architectures \cite{sarti2023inseq}. These methods allow researchers to study whether model predictions are based on meaningful evidence or shortcut cues.

\subsection{Calibration and model reliability}
Model calibration measures the alignment between predicted confidence and empirical accuracy, and is commonly used as a proxy for reliability in neural models. However, recent work has shown that calibration metrics alone may not reflect the true robustness of model decisions. In particular, models with low calibration error can still rely on non-generalizable features, leading to the reliability paradox, where statistically well-calibrated models make unreliable decisions \cite{bihani2024reliability}. This suggests that evaluating reliability requires analyzing both confidence estimates and the underlying decision rules.

\subsection{Reliability and shortcut learning in large language models}
Recent work has begun to investigate whether large language models overcome shortcut learning. While scaling improves performance, studies show that LLMs can still rely on spurious correlations and dataset-specific cues, especially in structured evaluation settings \cite{yuan2024llms}. These findings motivate the need for evaluation methods that jointly analyze calibration, attribution, and robustness in modern decoder-only models.

% \subsection{Gap and our contribution}
Prior work on the reliability paradox focuses on encoder-based classification models and does not consider the setting of machine unlearning, where model parameters are explicitly modified to remove specific training data. In the unlearning scenario, a model may maintain good calibration while internally changing its decision rules, potentially increasing reliance on shortcut cues. This behavior has not been systematically studied for large decoder-only language models.

In this work, we extend the analysis of calibration and shortcut learning to the unlearning setting. Using the RELU evaluation protocol on the TOFU benchmark, we evaluate reliability of pretrained, fully fine-tuned, and unlearned models using both calibration metrics and attribution-based shortcut detection. Our results show that well-calibrated models after unlearning can still rely on correlation-based tokens, demonstrating that the reliability paradox also appears in unlearned language models.

% \section{Example Appendix}
\section{Calibration Analysis Across Forget Ratios}
\label{app:calibration_curves}

In this section, we present reliability diagrams for models evaluated under different forget ratios (1\%, 5\%, and 10\%) in the RELU-MCQA setting. These plots visualize the relationship between model confidence and empirical accuracy, providing insight into probabilistic reliability.

Each figure shows calibration curves for multiple model variants, including pretrained, fully fine-tuned, retained, and unlearned models (e.g., GradAscent, GradDiff, NPO, DPO). The diagonal line represents perfect calibration, where predicted confidence matches observed accuracy. Deviations from this line indicate miscalibration.

Across all forget ratios, we observe that fine-tuned and unlearned models often remain well calibrated, with confidence closely tracking accuracy. However, as shown in the main paper, this apparent calibration does not necessarily reflect reliable decision-making, as these models may still rely on shortcut cues.

\begin{figure*}[ht!]
    \centering
    \includegraphics[width=\textwidth]{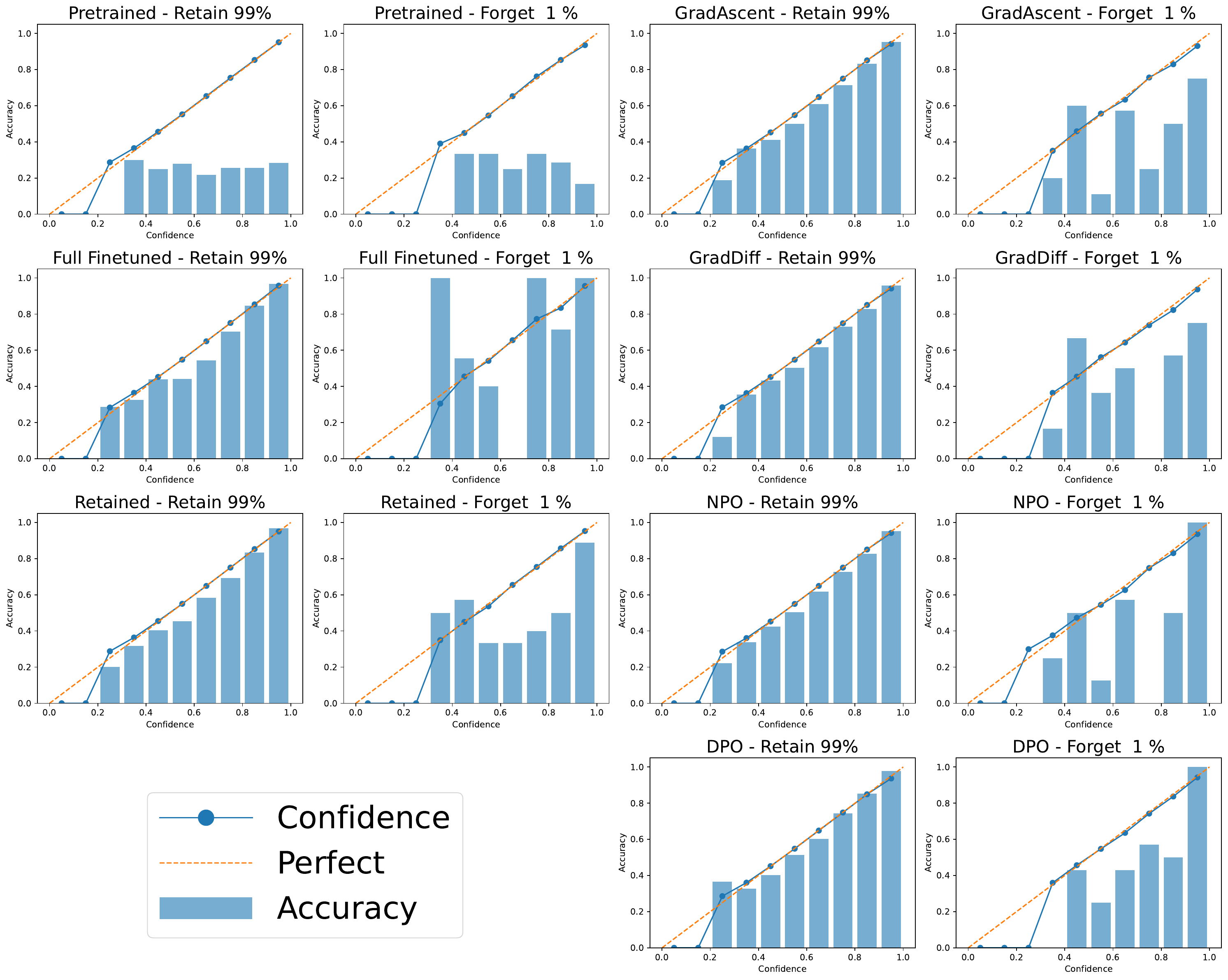}
    \caption[Calibration curves for 1\% forget setting.]{
    \textbf{Calibration curves for 1\% forget setting.}
    Reliability diagrams for pretrained, fully fine-tuned, retained, and unlearned models under 1\% forgetting. The diagonal line indicates perfect calibration. While fine-tuned and unlearned models exhibit low calibration error, several models show deviations in mid-confidence regions, suggesting imperfect alignment between confidence and accuracy.
    }
    \label{fig:calibration_1}
\end{figure*}

\begin{figure*}[ht!]
    \centering
    \includegraphics[width=\textwidth]{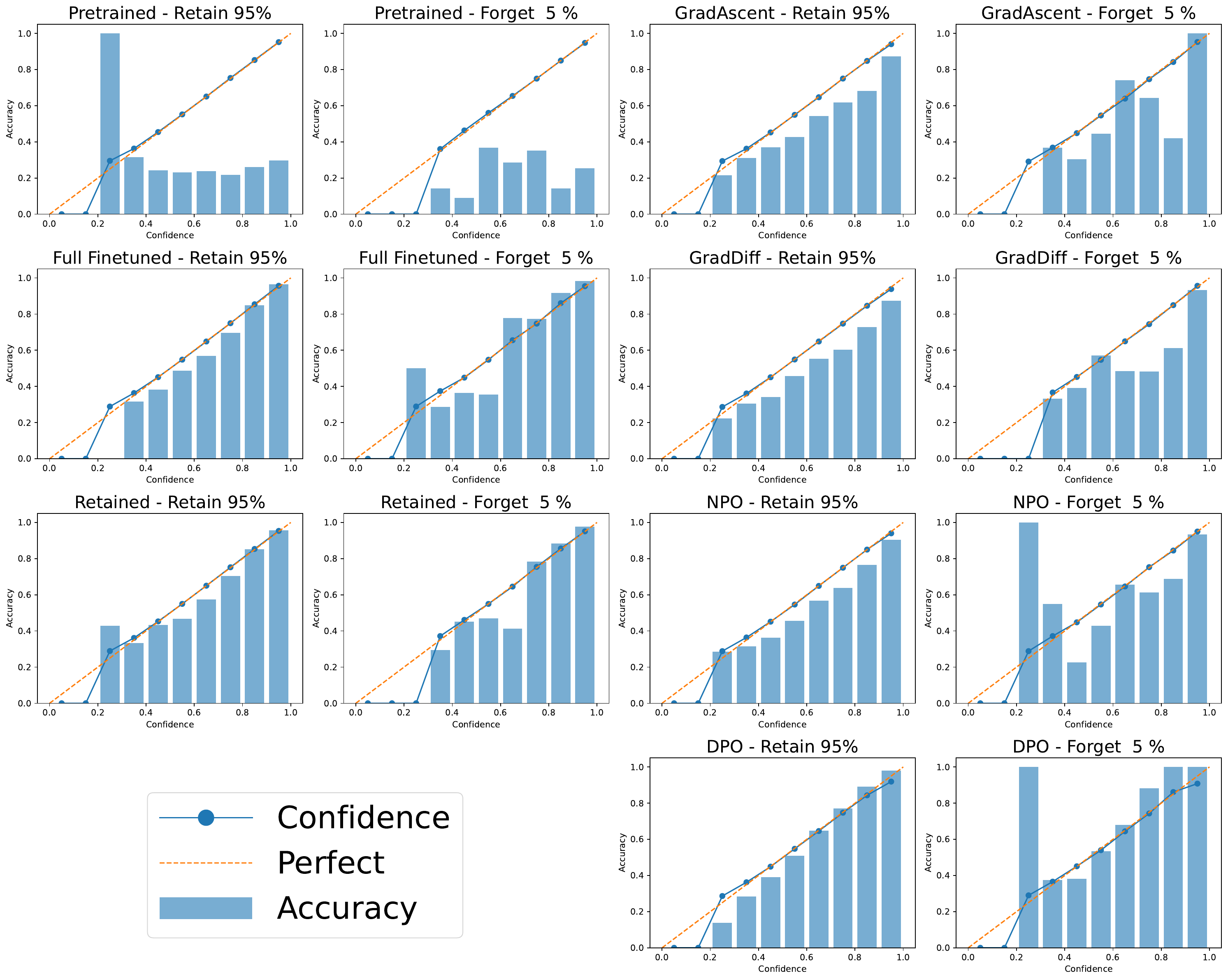}
    \caption[Calibration curves for 5\% forget setting.]{
    \textbf{Calibration curves for 5\% forget setting.}
    Reliability diagrams under increased forgetting. Models maintain relatively good calibration overall, with confidence closely tracking accuracy. However, variability across models increases compared to the 1\% setting, particularly in intermediate confidence bins.
    }
    \label{fig:calibration_5}
\end{figure*}

Figure~\ref{fig:calibration_1} shows results for 1\% forgetting (see :contentReference[oaicite:0]{index=0}), while Figures~\ref{fig:calibration_5} and~\ref{fig:calibration_10} present results for higher forget ratios.

\section{Calibration Analysis across different bin sizes and forget ratios} 
\label{app:calibration-bin-sizes}

Both Expected Calibration Error (ECE) and Maximum Calibration Error (MCE) rely on a binning approach: predictions are first partitioned into bins based on confidence, and the gap between accuracy and average confidence is then aggregated within each bin. 
Consequently, the number of bins $n$ directly influences the reported error---too few bins coarsen genuine miscalibration,  while too many introduce sample-size bias as bins become sparsely populated. 
In this section, we examine the sensitivity of ECE and MCE to bin size  ($n \in \{10, 20, 33, 100\}$) across all forget ratios.

Table~\ref{tab:results-bins-compare} shows that on retain splits (99\%, 95\%, 90\%), ECE values remain largely unchanged across $n=10$, $20$, and $33$, with only a marginal increase at $n=100$. 
For instance, the fully finetuned model on retain 90\% reports ECE values of 0.039, 0.040, 0.041, and 0.051 across the four bin sizes. This stability is expected because the retain splits contain enough samples that even fine-grained binning yields reasonably populated buckets, keeping the binned ECE bias low. 
In contrast, ECE values on forget splits increase consistently with bin size. For example, on forget 1\%, the fully finetuned model's ECE rises from 0.174 ($n=10$) to 0.298 ($n=100$), and similar trends hold across all unlearning methods. 
We attribute this to the smaller size of the forget splits: as $n$ grows, individual bins contain fewer datapoints, increasing both the variance and the bias of ECE. 
This is a well-known artifact of binned calibration estimators on small samples rather than a genuine change in model calibration \citep{pavlovic2025understandingmodelcalibration}.

MCE shows more volatility across bin sizes than ECE, since a single 
sparsely populated bin can dominate the maximum. On most retain splits MCE 
increases monotonically with $n$, but the retained model on retain 95\% 
deviates from this trend as the MCE jumps from 0.140 at $n=10$ to 0.729 at $n=33$, then drops to 0.464 at $n=100$. 
This non-monotonic behavior reflects the worst-case nature of MCE: as bins are redistributed, the bin containing the largest accuracy--confidence gap can shift abruptly, making MCE a less stable summary statistic when bin populations are uneven.

We adopt $n=10$ for the main results for two reasons. First, 10 bins 
yield enough datapoints per bucket on both retain and forget splits to produce a low-bias ECE estimate and is popular in previous studies on calibration estimation \citep{guo2017calibration, pavlovic2025understandingmodelcalibration}. 
Second, we observe that the primary finding on reliability from the lens of calibration and shortcut usage is preserved across $n=10$, 
$20$, and $33$ on the retain splits, indicating that the \textit{reliability paradox} is not an artifact of a particular bin choice. 
These observations indicate that the calibration error of model before and after unlearning on retain splits is a robust property, irrespective of the bin size.

\begin{table*}[t]
\centering
\resizebox{\textwidth}{!}{
\begin{tabular}{lcccc|cccc|cccc|cccc}
\toprule
 & \multicolumn{8}{c|}{\textbf{Forget Split}} & \multicolumn{8}{c}{\textbf{Retain Split}} \\
\cmidrule(lr){2-9} \cmidrule(lr){10-17}
 & \multicolumn{4}{c|}{ECE ($\downarrow$)} & \multicolumn{4}{c|}{MCE ($\downarrow$)} & \multicolumn{4}{c|}{ECE ($\downarrow$)} & \multicolumn{4}{c}{MCE ($\downarrow$)} \\
\cmidrule(lr){2-5} \cmidrule(lr){6-9} \cmidrule(lr){10-13} \cmidrule(lr){14-17}
\textbf{Model} & $n{=}10$ & $n{=}20$ & $n{=}33$ & $n{=}100$ & $n{=}10$ & $n{=}20$ & $n{=}33$ & $n{=}100$ & $n{=}10$ & $n{=}20$ & $n{=}33$ & $n{=}100$ & $n{=}10$ & $n{=}20$ & $n{=}33$ & $n{=}100$ \\
\midrule

\multicolumn{17}{c}{\textbf{Forget 1\% / Retain 99\%}} \\
\midrule
Pretrained & 0.435 & 0.493 & 0.488 & 0.543 & 0.768 & 0.930 & 0.929 & 0.943 & 0.519 & 0.521 & 0.521 & 0.521 & 0.668 & 0.674 & 0.674 & 0.710 \\
Full       & 0.174 & 0.204 & 0.247 & 0.298 & 0.695 & 0.695 & 0.695 & 0.857 & 0.040 & 0.045 & 0.046 & 0.058 & 0.107 & 0.119 & 0.733 & 0.733 \\
Retained   & 0.212 & 0.249 & 0.286 & 0.384 & 0.357 & 0.827 & 0.827 & 0.871 & 0.047 & 0.047 & 0.047 & 0.055 & 0.096 & 0.119 & 0.270 & 0.297 \\
GradAscent & 0.266 & 0.344 & 0.317 & 0.423 & 0.506 & 0.738 & 0.738 & 0.843 & 0.032 & 0.035 & 0.038 & 0.054 & 0.097 & 0.097 & 0.109 & 0.260 \\
GradDiff   & 0.239 & 0.333 & 0.309 & 0.418 & 0.739 & 0.786 & 0.786 & 0.844 & 0.026 & 0.032 & 0.038 & 0.052 & 0.164 & 0.164 & 0.130 & 0.285 \\
NPO        & 0.245 & 0.279 & 0.331 & 0.398 & 0.748 & 0.894 & 0.774 & 0.894 & 0.029 & 0.033 & 0.040 & 0.056 & 0.064 & 0.075 & 0.094 & 0.392 \\
DPO        & 0.212 & 0.223 & 0.259 & 0.412 & 0.360 & 0.525 & 0.533 & 0.818 & 0.033 & 0.036 & 0.036 & 0.048 & 0.081 & 0.081 & 0.269 & 0.341 \\

\midrule
\multicolumn{17}{c}{\textbf{Forget 5\% / Retain 95\%}} \\
\midrule
Pretrained & 0.518 & 0.525 & 0.522 & 0.541 & 0.707 & 0.730 & 0.811 & 0.993 & 0.523 & 0.523 & 0.523 & 0.525 & 0.705 & 0.705 & 0.705 & 0.721 \\
Full       & 0.082 & 0.121 & 0.107 & 0.176 & 0.211 & 0.331 & 0.668 & 0.715 & 0.040 & 0.042 & 0.046 & 0.049 & 0.289 & 0.289 & 0.266 & 0.297 \\
Retained   & 0.067 & 0.089 & 0.138 & 0.207 & 0.232 & 0.416 & 0.497 & 0.825 & 0.032 & 0.036 & 0.037 & 0.052 & 0.140 & 0.140 & 0.729 & 0.464 \\
GradAscent & 0.129 & 0.147 & 0.132 & 0.213 & 0.421 & 0.535 & 0.578 & 0.874 & 0.108 & 0.108 & 0.109 & 0.111 & 0.165 & 0.166 & 0.199 & 0.286 \\
GradDiff   & 0.109 & 0.142 & 0.149 & 0.226 & 0.263 & 0.424 & 0.569 & 0.896 & 0.100 & 0.101 & 0.101 & 0.107 & 0.145 & 0.149 & 0.738 & 0.738 \\
NPO        & 0.128 & 0.149 & 0.126 & 0.211 & 0.712 & 0.712 & 0.712 & 0.767 & 0.083 & 0.083 & 0.084 & 0.087 & 0.112 & 0.146 & 0.260 & 0.260 \\
DPO        & 0.052 & 0.086 & 0.124 & 0.199 & 0.710 & 0.710 & 0.710 & 0.710 & 0.046 & 0.047 & 0.051 & 0.061 & 0.149 & 0.149 & 0.232 & 0.400 \\

\midrule
\multicolumn{17}{c}{\textbf{Forget 10\% / Retain 90\%}} \\
\midrule
Pretrained & 0.497 & 0.497 & 0.499 & 0.511 & 0.662 & 0.676 & 0.724 & 0.825 & 0.516 & 0.516 & 0.516 & 0.519 & 0.665 & 0.668 & 0.695 & 0.738 \\
Full       & 0.068 & 0.075 & 0.089 & 0.136 & 0.273 & 0.273 & 0.273 & 0.827 & 0.039 & 0.040 & 0.041 & 0.051 & 0.070 & 0.083 & 0.272 & 0.272 \\
Retained   & 0.083 & 0.109 & 0.125 & 0.168 & 0.298 & 0.332 & 0.327 & 0.620 & 0.038 & 0.039 & 0.041 & 0.057 & 0.081 & 0.092 & 0.112 & 0.305 \\
GradAscent & 0.245 & 0.245 & 0.248 & 0.252 & 0.707 & 0.707 & 0.707 & 0.707 & 0.209 & 0.209 & 0.209 & 0.209 & 0.383 & 0.495 & 0.557 & 0.696 \\
GradDiff   & 0.136 & 0.136 & 0.142 & 0.184 & 0.285 & 0.388 & 0.431 & 0.905 & 0.133 & 0.133 & 0.134 & 0.136 & 0.221 & 0.221 & 0.268 & 0.343 \\
NPO        & 0.108 & 0.115 & 0.120 & 0.165 & 0.213 & 0.258 & 0.342 & 0.719 & 0.101 & 0.101 & 0.107 & 0.111 & 0.159 & 0.191 & 0.180 & 0.239 \\
DPO        & 0.046 & 0.057 & 0.103 & 0.139 & 0.099 & 0.140 & 0.227 & 0.501 & 0.052 & 0.052 & 0.054 & 0.062 & 0.089 & 0.090 & 0.263 & 0.276 \\

\bottomrule
\end{tabular}
}
\caption[ECE and MCE for \texttt{Llama-3.1-8B} across different forget ratios.]{ECE and MCE for \texttt{Llama-3.1-8B} across forget ratios 1\%, 5\%, and 10\%, and bin sizes ranging from 10 to 100. Forget and retain splits are reported side-by-side to enable direct comparison of calibration error behavior on data targeted for unlearning versus data to be preserved.}
\label{tab:results-bins-compare}
\end{table*}

\section{Qualitative Examples} \label{app:qualitative-examples}
In Table~\ref{tab:shortcut_tokens} we present some qualitative examples from a model unlearned using DPO on a 90\% retain split ratio. We observe that the model relies on specific words, not pertaining to the author's identity or the specific question, to make decisions. For example, in the first question, the model relies on ``been'' to make a decision. Similar things are also observed in the remaining qualitative examples. A more comprehensive list of model attributions, top tokens identified by LMI, and identified shortcuts is present in our GitHub repo \url{https://github.com/Exploration-Lab/Unlearning-Reliability-Paradox}.

\begin{table*}[ht]
\centering
\footnotesize
\setlength{\tabcolsep}{4pt}
\renewcommand{\arraystretch}{1.25}
\begin{tabularx}{\textwidth}{@{} p{0.4cm} L C{1.8cm} C{1.3cm} p{4.2cm} @{}}
\toprule
\textbf{\#} & \textbf{Question \& Options} & \textbf{Generated Answer} & \textbf{Ground-truth} & \textbf{Shortcut Tokens} \\
\midrule
 
% --- Question 1 ---
1 & Has Alejandro Cordero Rodriguez \hl{been} the recipient of any prestigious awards?\newline
\textbf{A.} The Silver Scroll Award for Fantasy Fiction \newline
\textbf{B.} The Bronze Book Award for Mystery Novels \newline
\textbf{C.} The Golden Pen Award for Science Fiction \newline
\textbf{D.} The Sapphire Quill Award for Alternate History
& D & D &
\begin{tabular}{@{}lll@{}}
\textit{Token} & \textit{Attr.} & \textit{LMI} \\
\hline
\texttt{been} & 0.0112 & 0.0108 \\
\end{tabular} \\
\midrule
 
% --- Question 2 ---
2 & How \hl{does} Andres Santiago Cruz's family background influence his writing \hl{about} parenting? \newline
\textbf{A.} Raised by a Conservationist and a Veterinarian, Andres Santiago Cruz's writings often include ideas about nurturing and caring, drawn from his formative experiences. \newline
\textbf{B.} Raised by a Lawyer and a Politician, Andres Santiago Cruz's writings often include ideas about justice and governance. \newline
\textbf{C.} Raised by a Chef and a Musician, Andres Santiago Cruz's writings often include ideas about creativity and expression. \newline
\textbf{D.} Raised by a Scientist and an Engineer, Andres Santiago Cruz's writings often include ideas about innovation and problem-solving.
& A & A &
\begin{tabular}{@{}lll@{}}
\textit{Token} & \textit{Attr.} & \textit{LMI} \\
\hline
\texttt{does}  & 0.0140 & 0.0578 \\
\texttt{about} & 0.0103 & 0.0543 \\
\end{tabular} \\
\midrule
 
% --- Question 3 ---
3 & How \hl{does} Catherine Marianne Pfeiffer's upbringing feature \hl{in} her writing? \newline
\textbf{A.} Influenced by her parents' careers in science and technology, Catherine Marianne Pfeiffer's narratives often revolve around innovation and progress. \newline
\textbf{B.} Influenced by her parents' careers in medicine and education, Catherine Marianne Pfeiffer's narratives often revolve around health and learning. \newline
\textbf{C.} Influenced by her parents' careers in finance and law, Catherine Marianne Pfeiffer's narratives often revolve around money and justice. \newline
\textbf{D.} Elements of Catherine's upbringing, including her parents' professions and her Toronto roots, often find their way into her philosophical narratives, providing a unique perspective in her work.
& D & D &
\begin{tabular}{@{}lll@{}}
\textit{Token} & \textit{Attr.} & \textit{LMI} \\
\hline
\texttt{does} & 0.0078 & 0.0415 \\
\texttt{in}   & 0.0066 & 0.0130 \\
\end{tabular} \\
\midrule
 
% --- Question 4 ---
4 & Why is Yigal Abramovitz known \hl{for} his humor literature? \newline
\textbf{A.} Yigal Abramovitz is known \hl{for} his intense psychological thrillers. \newline
\textbf{B.} Yigal Abramovitz is known \hl{for} his dark and brooding poetry. \newline
\textbf{C.} Yigal Abramovitz is known \hl{for} his slapstick comedy in his novels. \newline
\textbf{D.} Yigal Abramovitz had a unique way to infuse humor in everyday-life situations, making them relatable yet hilarious, which became his trademark.
& D & D &
\begin{tabular}{@{}lll@{}}
\textit{Token} & \textit{Attr.} & \textit{LMI} \\
\hline
\texttt{for} & 0.0195 & 0.0025 \\
% \texttt{for} & 0.0165 & 0.0025 \\
% \texttt{for} & 0.0116 & 0.0025 \\
\end{tabular} \\
\bottomrule
\end{tabularx}
\caption[Qualitative examples showing shortcut tokens]{Multiple-choice questions with model-selected answers, ground-truth answers, and the shortcut tokens identified during attribution analysis. Each shortcut token row reports the token string, its attribution, and its LMI (Lexical Mutual Information) score. The identified token is \hl{highlighted in the text}. These examples are extracted from the unlearned model produced by DPO on the retain 90\% split. More qualitative examples are present on our github repository.}
\label{tab:shortcut_tokens}
\end{table*}

\end{document}